\documentclass[11pt,a4paper]{article}
\usepackage{emnlp-ijcnlp-2019}
\usepackage{times}
\usepackage{latexsym}

\usepackage{natbib}
\usepackage{url}
\usepackage{verbatim}
\usepackage{subcaption}
\usepackage{graphicx}
\usepackage{rotating}
\usepackage{supertabular}
\usepackage{longtable}
\usepackage{amsmath}
\usepackage{amssymb}

\aclfinalcopy % Uncomment this line for the final submission

\title{When to reply? Context Sensitive Models to Predict 
Instructor Interventions in MOOC Forums}

\author{Muthu Kumar Chandrasekaran$^{1,2}$ 
  Min-Yen Kan$^1$\\
$^1$Dept. of Computer Science 
  National University of Singapore
  Singapore \\
$^2$SRI International 
  Menlo Park CA, USA \\
  {\tt cmkumar087@gmail.com}\\
  {\tt knmnyn@comp.nus.edu.sg}\\}

\date{}

\begin{document}
\maketitle
\begin{abstract}
  %background sentence
 Due to time constraints, course instructors often need to 
 selectively participate in student discussion threads, due 
 to their limited bandwidth and 
 lopsided student--instructor ratio on online 
 forums. 
 % Earlier works have proposed to predict threads 
 % that the instructor(s) should intervene as a binary classification. 
 We propose the first deep learning models for this  
 binary prediction problem. 
 We propose novel attention based 
 models to infer the amount of latent context necessary to predict  
 instructor intervention. Such models also allow themselves to be tuned 
 to instructor's preference to 
 intervene early % (e.g., in a STEM MOOC) or intervene late
 or late.
% (e.g., in a Humanities MOOC) in student discussions. 
 Our four proposed 
 attentive model variants % to infer the latent context 
 improve over 
 the state-of-the-art by a significant, large margin of 11\% 
 in $F1$ and 10\% in 
 recall, on average. Further, introspection of attention help us better understand 
 what aspects of a discussion post propagate through the discussion thread that prompts instructor intervention.
\end{abstract}

\section{Introduction}
\label{s:intro}
Massive Open Online Courses (MOOCs) have strived to bridge the social gap in 
higher education by bringing quality education from reputed universities 
to students at large. 
Such massive scaling through online classrooms, 
however, disrupt co-located, synchronous two-way 
communication between the students and the instructor. 
\begin{figure}[ht]
\includegraphics[width=0.5\textwidth]{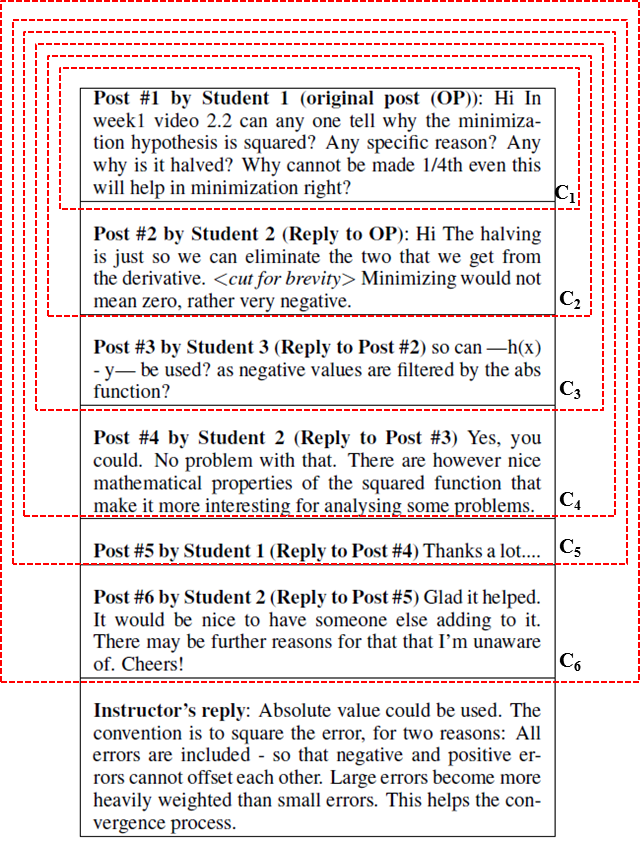}
\caption{An example discussion thread from the ml-005 MOOC where 
latent context candidates are marked with dotted red lines for 
instructor's reply.}
\label{fig:example}
\end{figure}
MOOC platforms provide discussion forums for students to 
talk to their classmates about the lectures, homeworks, quizzes and 
provide a venue to socialise. Instructors (defined here as the course instructors, their teaching assistants 
and the MOOC platform's technical staff) monitor the discussion forum to 
post (reply to their message) in discussion threads among students. We 
refer to this posting as {\it intervention}, following prior work~\cite{chaturvedi14}.
However, due to large student enrolment, the student--instructor ratio in MOOCs
% Min NAACL
% is very low.   (<-student/faculty not faculty/student as you designated.  Otherwise reverse here and in abstract) 
is very high
Therefore, instructors are not able to monitor and participate in 
all student discussions. To address this problem, a number of works have proposed 
systems e.g.,~\cite{chaturvedi14,chandrasekaran2017using} to aid instructors to 
selectively intervene on student discussions where they are needed the most. 

\begin{figure}
% Min NAACL give more room to figure between the dashes and lighten them.  They are too heavy.
    \centering
    \includegraphics[width=0.5\textwidth]{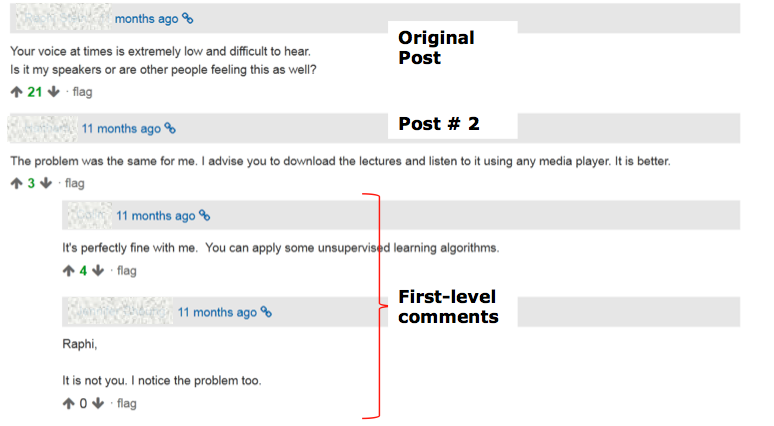}
    \caption{Structure of the discussion threads on Coursera with posts 
    and one level of comments. Our models are aimed at capturing the structure 
    at this level.}
    \label{fig:thread_struct}
\end{figure}

In this paper, we improve the state-of-the-art for instructor intervention in MOOC 
forums. We propose the first neural models for this prediction problem. We show 
that modelling the thread structure and the sequence of posts explicitly improves 
performance. Instructors in different MOOCs from different subject areas intervene 
differently. For example, on a Science, Technology, Engineering and Mathematics (STEM) 
MOOC, instructors may often intervene early as possible 
% Min NAACL do you have a citation for this claim?  Otherwise, hedge.
to resolve misunderstanding of the subject material and prevent confusion. However, 
in a Humanities MOOC, instructors allow for the students to explore open-ended 
discussions and debate among themselves. 
% Min: NAACL BUG these two cases conflic (later/earlier)
Such instructors may prefer to intervene 
later in the discussion to encourage further discussion or resolve conflicts among 
students. We therefore propose attention models to infer the latent \textit{context}, 
i.e., the series of posts that trigger an intervention. 
% Min NAACL : studies on MOOCs?  Or in general for discussion contexts?
Earlier studies on MOOC forum intervention either model the 
entire context or require the context size to be specified explicitly.  

\subsection{Problem Statement}
\label{ss:prob_stmt}
A thread $T$ consists of a series of posts $P_1$ through $P_n$ where 
% $P_n$ is an instructor's post when $T$ is intervened., if applicable.  
% Min NAACL BUG what if not intervened?
$P_n$ is an instructor's post when $T$ is intervened, if applicable.  
$T$ is considered intervened if an instructor had posted at least once.
% Min: overly verbose.
The problem of predicting instructor intervention is cast as a binary 
classification problem. Intervened threads are predicted as 1 given 
while non-intervened threads are predicted as 0 given posts $P_1$ 
through $P_{n-1}$.

% Min: NAACL The problem here is that you have a secondary problem of defining the appropriate amount of context.  Do you want to phrase it directly as such?
The primary problem leads to a secondary problem of 
inferring the appropriate amount of context to intervene. 
We define a context $C_i$ of a post $P_i$ as a series of linear 
contiguous posts $P_1$ through $P_j$ where $j<=i$. The problem of 
inferring context is to identify context $C_i$ from a set of 
candidate contexts ${C_1,\dots,C_{n-1}}$.

%\textbf{Background on Instructor intervention research.}  Should instructors intervene in 
%student discussions in face-to-face or virtual classrooms is an open problem 
%in pedagogy~\cite{mazzolini2003}. Further, when to intervene~\cite{mazzolini2007} 
%in such discussions is natural follow-up research question that remains to be 
%investigated. Prior results and indicative studies have polarized the pedagogy 
%research community to lean towards one or the other argument. MOOCs offer an 
%opportunity to test these hypotheses definitively due to its scale that can render 
%significance to the results. In this paper, we seek to address this research gap.

%%%%%%%%%%%%%%%%%%%%%%%%%%%%%
\section{Related Work}
\label{s:related}
\subsection{Modelling Context in Forums}
\label{ss:}
\label{ss:rel_context}
Context has been used and modelled in various ways for different problems 
in discussion forums. In a work on a closely related problem of forum thread 
retrieval~\citet{wang2013utility} models 
context using inter-post discourse e.g., Question-Answer.
\citet{wang2011learning} models the structural 
dependencies and relationships between forum posts using a conditional random field
in their problem to infer the reply structure. Unlike 
\citet{wang2013utility}, \citet{wang2011learning} can be used to model any structural dependency and is, therefore, more general. 
In this paper, we seek to infer general dependencies between a reply 
and its previous context whereas \cite{wang2011learning} inference 
is limited to pairs of posts.
More recently~\cite{jiao2018} proposed a context based model 
which factorises attention over threads of different lengths. Differently, 
we do not model length but the context before a post. However, our 
attention models cater to threads of all lengths.

\citet{zayats2018conversation}~proposed graph structured LSTM to model the explicit reply structure in Reddit forums.
Our work does not assume access to such a reply structure because 1) Coursera 
forums do not provide one and 2) forum participants often err by posting their 
reply to a different post than that they intended. At the other end of the 
spectrum are document classification models that do not assume structure in 
the document layout but try to infer inherent structure in the natural language, 
viz, words, sentences, paragraphs and documents. Hierarchical 
attention~\cite{yang2016hierarchical} is a well know recent work that 
classifies documents using a multi-level LSTMs with attention mechanism 
to select important units at each hierarchical level. Differently, we 
propose a hierarchical model that encodes layout hierarchy between a post 
and a thread but also infers reply structure using a attention mechanism 
since the layout does not reliably encode it.

\subsection{Instructor Intervention in MOOC forums}
\label{ss:rel_intervention}
The problem of predicting instructor intervention in 
MOOCs was proposed by~\cite{chaturvedi14}. 
Later~\citet{chandrasekaran2015learning} 
evaluated baseline models by \cite{chaturvedi14} 
over a larger corpus and found the results to vary widely 
across MOOCs. Since then subsequent works 
have used similar diverse evaluations on the 
same prediction problem~\cite{chandrasekaran2017using,chandrasekaran2018countering}. 
\citet{chandrasekaran2017using} proposed models with discourse features to enable better prediction over unseen 
MOOCs. \citet{chandrasekaran2018countering} 
recently showed interventions on Coursera forums to be 
biased by the position at which a thread appears to an 
instructor viewing the forum interface and proposed methods 
for debiased prediction. 

While all works since~\citet{chaturvedi14} address key limitations 
in this line of research, they have not investigated the role  
of structure and sequence in the threaded discussion in 
predicting instructor interventions. 
\citet{chaturvedi14} proposed probabilistic graphical models 
to model structure and sequence. They inferred vocabulary 
dependent latent post categories to model the thread sequence 
and infer states that triggered intervention. Their model, 
however, requires a hyperparameter for the number of latent 
states. It is likely that their empirically reported 
setting will not generalise due to their weak evaluation~\cite{chandrasekaran2015learning}. 
In this paper, we propose models to infer the context that triggers 
instructor intervention that does not require context lengths 
to be set apriori. All our proposed models generalise over modelling 
assumptions made by~\citet{chaturvedi14}.

For the purpose of comparison against a state-of-the-art and competing 
baselines we choose \cite{chandrasekaran2015learning} since 
\cite{chaturvedi14}'s system and data are not available for replication.

\section{Data and Preprocessing}
\label{s:data}
We evaluate our proposed models over a 
corpus of 12 MOOC iterations (offerings) on 
    Coursera.org\footnote{Coursera is a commercial MOOC platform accessible 
    at \url{https://www.coursera.org}}
In partnership with Coursera and in line 
with its Terms of Service, we obtained the data for 
use in our academic research. Following prior work~\cite{chandrasekaran2015learning} 
we evaluate over a diverse dataset to represent 
MOOCs of varying sizes, instructor styles, instructor 
team sizes and number of threads intervened.
We only include threads from sub-forums on Lecture, Homework, 
Quiz and Exam. We also normalise and label sub-forums with 
other non-standard names (e.g., Assignments 
instead of Homework) into of the four said sub-forums.
Threads on general discussion, meet and greet and other custom 
sub-forums for social chitchat are omitted as our focus is to aid 
instructors on intervening on discussion on the subject matter. 
We also exclude announcement threads and other threads started 
by instructors since they are not interventions. We preprocess 
each thread by replacing URLs, equations and other mathematical 
formulae and references to timestamps in lecture videos by tokens 
$<$URL$>$, $<$MATH$>$, $<$TIMEREF$>$ respectively. We also truncate 
intervened threads to only include posts before the first 
instructor post since the instructor's and subsequent posts will 
bias the prediction due to the instructor's post.

%Threads that have been replied to at least once 
%by an instructor, a teaching assistant or a community teaching assistant (hereafter, 
%referred to as \textit{instructor}) as an \textit{intervention} in line with prior %research~\cite{chandrasekaran2015learning,chaturvedi14}.

%%%%%%%%%%%%%%%%%%%%%%%%%%%%%%%%%%%%%%%%%%%%%%%%%%%%%%%%%%%%%%%%%%%%%
\section{Model}
\label{s:model}

The key innovation of our work is to decompose the intervention prediction problem into a two-stage model that first explicitly tries to discover the proper context to which a potential intervention could be replying to, and then, predict the intervention status.
This model implicitly assesses the importance (or urgency) of the existing thread's context to decide whether an intervention is necessary. For example in Figure~\ref{s:intro}, prior to the instructor's intervention, the ultimate post (Post \#6) by Student~2 already acknowledged the OP's gratitude for his answer.  In this regard, the instructor may have decided to use this point to summarize the entire thread to consolidate all the pertinent positions.  Here, we might assume that the instructor's reply takes the entire thread (Posts \#1--6) as the context for her reply.

This subproblem of inferring the context scope is where our innovation centers on.  To be clear, in order to make the prediction that a instruction intervention is now necessary on a thread, the instructor's reply is not yet available --- the model predicts whether a reply is necessary --- so in the example, only Posts \#1--6 are available in the problem setting.  To infer the context, we have to decide which subsequence of posts are the most plausible motivation for an intervention.

Recent work in deep neural modeling has used an attention mechanism as a focusing query to highlight specific items within the input history that significantly influence the current decision point.  Our work employs this mechanism -- but with a twist: due to the fact that the actual instructor intervention is not (yet) available at the decision timing, we cannot use any actual intervention to decide the context.  To employ attention, we must then employ a surrogate text as the query to train our prediction model.  Our model variants model assess the suitability of such surrogate texts for the attention mechanism basis. 

Congruent with the representation of the input forums, in all our proposed models, we encode the discussion thread hierarchically. 
We first build representations for each post by passing pre-trained word vector 
representations from GloVe~\cite{pennington2014glove} for each word through 
an LSTM~\cite{hochreiter1997long}, $lstm_{post}$. We use the last layer output of the 
LSTM as a representation of the post. We refer this as the post vector $P_i$.

\begin{figure}[ht]
\centering
\includegraphics[]{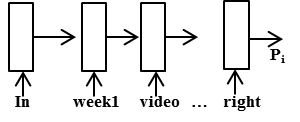}
% Min NAACL repetitive caption.  Fix if have time.
\caption{An LSTM post encoder whose last layer output 
is taken as the post representation or post vector $P_i$.}
\label{fig:postenc}
\end{figure}

Then each post $P_i$ is passed through another 
LSTM, $lstm_{ctx}$, whose last layer  
output forms the encoding of the entire thread. 
Hidden unit outputs of $lstm_{ctx}$ represent  
the contexts $C$; that is, snapshots of the threads after each 
post, as shown in Figure~\ref{fig:example}.

The $lstm_{post}$ and $lstm_{ctx}$ together constitute the 
hierarchical LSTM (hLSTM) model.  This general hLSTM model serves 
as the basis for our model exploration in the rest of this section.
% Min NAACL do you have a citation for hLSTM?

%\begin{figure}[ht]
%\centering
%\includegraphics{context_encoder.png}
%\caption{LSTM context encoder where whose last layer output 
%is taken as the post representation or post vector $P_i$}
%\label{fig:example}
%\end{figure}

\subsection{Contextual Attention Models}
\label{ss:ctxattn}
When they intervene, instructors either pay attention 
to a specific post or a series of posts, which trigger their 
reply. However, instructors rarely explicitly indicate to which 
post(s) their intervention is in relation to. This is the case 
in our corpus, party due to Coursera's user interface 
which only allows for single level comments (see 
Figure~\ref{fig:thread_struct}).
% Min: NAACL hard to parse garden path sentence.  Rephrase
% So, based only on the thread-level 1 (intervened) or 0 (not intervened) 
%supervision signal we seek to infer the context, represented by a 
%sequence of posts, to which they intervene.
% Min: rephrased.  Please check.
Based solely on the binary, thread-level intervention signal, 
our secondary objective seeks to infer the appropriate 
context -- represented by a sequence of posts -- as the basis 
for the intervention. 

We only consider linear contiguous series of posts starting with 
the thread's original post to constitute to a context; e.g., 
$c_2=<p1,p2>$. This is a reasonable as MOOC forum posts always 
reply to the original post or to a subsequent post, which in 
turn replies to the original post. This is in contrast to forums 
such as Reddit that have a tree or graph-like structure that 
require forum structure to be modelled explicitly, such as 
in~\cite{zayats2018conversation}.

We propose three neural attention~\cite{bahdanau2014neural} variants 
based on how an instructor might attend and reply to a context in a 
thread: the ultimate, penultimate and any post attention 
models\footnote{\url{https://github.com/cmkumar87/Context-Net}}.
%\footnote{\url{https://github.com/cmkumar87/Context-Net}}
We review each of these in turn.~\\

\textbf{Ultimate Post Attention (UPA) Model.} In this model we 
attend to the context represented by hidden state of the 
$lstm_{ctx}$. We use the post prior 
to the instructor's reply as a query over the contexts $C$ 
to compute attention weights $\alpha$, which are then used to 
compute the attended context representation $attn\_ctx$ (recall 
again that the intervention text itself is not available for this 
purpose). This attention formulation makes an equivalence between 
the final $P_{n-1}$ post and the prospective intervention, using Post $P_{n-1}$ as the query for finding the appropriate context $C \in 
\{C_1, C_2, ..., C_{n-1}\}$, {\it inclusive} of itself $P_{n-1}$.  
Said in another way, UPA uses the most recent content in the thread 
as the attentional query for context.

\begin{figure}[ht]
\centering
\includegraphics{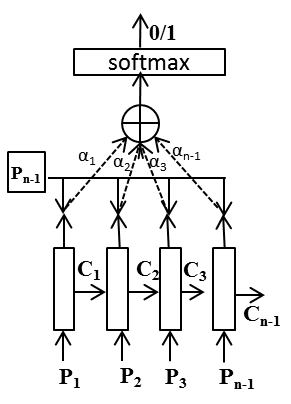}
\caption{Ultimate Post Attention model: The last post before the 
instructor's post is set as a 
query to attend over all the contexts prior to the post. 
% Min: Redundant. Push info back into the text where possible.
The model 
hypothesises that the instructor is  
likely reply to the same context that the post that immediately 
precedes the instructor's post,
% Min: omitted as I can't parse this.
% replied to, 
% Min: ??
since natural language conversations are typically 
structured that way. The post vector $P_i$ in this model are 
obtained from $lstm_{post}$, as shown in Figure~\ref{fig:postenc}.}
\label{fig:lastpostctx}
\end{figure}

% Min: better to have your example in the text below align with that in Figure 4.  Currently, it does not causing confusion.
For example, if post $P_3$ is the 
instructor's reply, post $P_2$ will query over the contexts 
$C_2=lstm_{ctx}(P_1,P_2)$ and $C_1=lstm_{ctx}(P_1)$.
The model schematic is shown in Figure~\ref{fig:lastpostctx}.

The attended context representations are computed as:
\begin{equation}
\label{eqn:upa}	
\begin{split}
    attn\_ctx = \sum_{i=1}^{n-1}{\alpha_i . c_i}\\
    \text{where, } \alpha_i = \frac{\exp{(a_i)}}{\sum_{i=1}^{n-1}{\exp{(a_i)}}}\\
           a_i = \mathbf{v}^\top \tanh(\mathbf{W}[P_{n-1};C_i]+\mathbf{b})
\end{split}
\end{equation}

The $attn\_ctx$ representation is then passed through a fully 
connected softmax layer to yield the binary prediction.~\\

\textbf{Penultimate Post Attention (PPA) Model.} 
% Min: I have no idea why you chose PPA to replace UPA.  It could have been any post.  So why compare with PPA?
% Min: ok will try to explain.
While the UPA model uses the most recent text and makes the ultimate 
post itself available as potential context, our
the ultimate post may be better modeled as having any 
of its prior posts as potential context.
Penultimate Post Attention (PPA) variant does this.
%Its attention formulation 
%models context directly, ignoring positional information.  
%
%It is conceivable 
%that an instructor might reply to a different context that is unconnected 
%to post that immediately precedes her intervention.  To realise one but likely 
%instance of this hypothesis, 
%
%We replace the query $P_{n-1}$ in the UPA model, 
%with $P_{n-2}$. 
The schematic and the equations for the PPA model are obtained 
by summing over contexts $c_1 \dots c_{n-2}$ in  
Equation~\ref{eqn:upa} and Figure~\ref{fig:lastpostctx}.
% Added by Min
While we could properly model such a context inference decision with 
any post $P_x$ and prospective contexts $C \in \{C_1,C_2,C_{x-1}\}$ 
(where $x$ is a random post), it makes sense to use the penultimate 
post, as we can make the most information available to the model for 
the context inference.  

The attended context representations are computed as:
\begin{equation}
\label{eqn:upa}	
\begin{split}
    attn\_ctx = \sum_{i=1}^{n-2}{\alpha_i . c_i}\\
    \text{where, } \alpha_i = \frac{\exp{(a_i)}}{\sum_{i=1}^{n-2}{\exp{(a_i)}}}\\
           a_i = \mathbf{v}^\top \tanh(\mathbf{W}[P_{n-1};C_i]+\mathbf{b})
\end{split}
\end{equation}

\textbf{Any Post Attention (APA) Model.} 
APA further relaxes both UPA and PPA, allowing APA to generalize and 
hypothesize that the prospective instructor intervention is based on 
the context that \textit{any previous post $P_i$} replied to.
In this model, each post $P_i$ is set 
as a query to attend to its previous context $C_{i-1}$. 
For example, $P_2$ will attend to $C_1$. Different from standard 
attention mechanisms,  APA attention 
weights $\alpha_i$ are obtained by normalising 
interaction matrix over the different queries.

\begin{figure}[ht]
\centering
\includegraphics[width=0.5\textwidth]{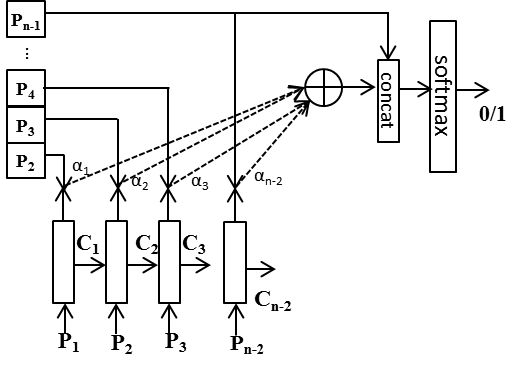}
\caption{The APA model is an extension of the UPA model in 
Figure~\ref{fig:lastpostctx}. In this model, each post $p_i$ is set 
as a query to attend to its previous context $c_{i-1}$, 
e.g., $p_2$ will attend to $c_1$
It hypothesises that the instructor may reply to a  
context that \textit{any previous post} replied to.}
\label{fig:anypostctx}
\end{figure}

In APA, the attention context $attn\_ctx$ is computed via:
\begin{equation}
\label{eqn:apa}
\begin{split}
    attn\_ctx = \sum_{i=1}^{n-2}{\alpha_i . c_i}\\
    \text{where } \alpha_i = \frac{\exp{(a_i)}}{\sum_{i=1}^{n-2}{\exp{(a_i)}}}\\
% Min: NAACL what's ``a''?  Do you mean ``a_i''?  If the same for all three models, why re-mention here?
a_i = \mathbf{v}^\top \tanh(\mathbf{W}[P_{i};C_{i-1}]+\mathbf{b}) \, .
\end{split}
\end{equation}

%%%%%%%%%%%%%%%%%%%%%%%%%%%%%%%%%%%%%%%%%%%%%%%%%%%%%%%%%%%%%%%%%%%%%
\section{Evaluation}
\label{s:eval}
% Min: you'll need to describe your partnership with Coursera and tell reviewers 
%that the discussions are within limits for the research with respect to IRB, 
%as prior papers have worried about this.
The baseline and the models are evaluated on a corpus of 12 MOOC 
discussion forums. We train on 80\% of 
the training data and report evaluation results on the held-out 20\% 
of test data. We report 
$F_1$ scores on the positive class (interventions), in line with 
prior work. We also argue that recall of the positive class is more 
important than precision, since it is 
costlier for instructors to miss intervening on a thread than 
spending irrelevant time intervening on a less critical threads 
due to false positives.

\textbf{Model hyperpameter settings.} All proposed and baseline 
neural models are trained using Adam optimizer with a learning rate 
of 0.001. 
% Min NAACL : Is this ok?  MB size == 1?? Muthu: yes MB size is 1
We used cross-entropy as loss function.
Importantly we updated the model parameters during 
training after each instance as in vanilla stochastic 
gradient descent; 
%this helped models to not overfit with a
this setting was practical since data on most courses had only a few 
hundred instances enabling convergence within a 
reasonable training time of a few hours (see 
Table~\ref{tab:result_attn_all}, column~2). 
% Min NAACL BUG: this and above looks very fishy to me.  Hope you can spend more time explaining.  These are very non-standard settings.
Models were trained for a single epoch as most of our courses with 
a few hundred thread converged after a single epoch.
% Min NAACL BUG: why dig out more exceptions?  Perhaps omit 
% Muthu: agree
% Training on ml-5 
% and rprog-3 converged after two epochs instead of one. 
% However, we choose to use the same 
% hyperpameter setting across all the courses to ensure that the results are comparable and easy to reproduce. 
% end possible omission
We used 300-dimensional GloVe vectors and permitted the embeddings 
to be updated during the model's end-to-end training.
% Min: why 128? %Muthu: hard to explain; empirical choice
The hidden dimension size of both $lstm_{post}$ and $lstm_{ctx}$ 
are set to 128 for all the models.

% Min: is your baseline open source which would explain your choice?  If so, say so.
%Muthu: yes, our systems are open source
% Min:
\textbf{Baselines.} We compare our models against a 
neural baseline models, hierarchical LSTM (hLSTM), 
with the attention ablated but with access to the complete context, 
and a strong, open-sourced feature-rich 
baseline~\cite{chandrasekaran2015learning}. 
We choose~\cite{chandrasekaran2015learning} over 
other prior works such as~\cite{chaturvedi14} since we 
do not have access to the dataset or 
the system used in their papers for replication.
\citet{chandrasekaran2015learning} 
is a logistic regression classifier with features inclusive of 
bag-of-words representation of the unigrams and thread length, 
normalised counts of agreements 
to previous posts, counts of non-lexical reference items such 
as URLs, and the Coursera forum type in which a thread appeared. 
We also report aggregated results from a hLSTM model with access 
only to the last post as context for 
comparison. Table~\ref{tab:baselines} compares the performance of 
these baselines against our proposed methods.  

\begin{table*}[ht]
    \centering
    \small
    \begin{tabular}{|l|r|r|r|r||r|r|r||r|r|r||r|r|r|}
    \hline
      \textbf{Course}& \textbf{\scriptsize Thread}& 
      \multicolumn{3}{|c||}{\textbf{hLSTM (Baseline)}}&
      \multicolumn{3}{|c||}{\textbf{UPA}}&
      \multicolumn{3}{|c||}{\textbf{PPA}}& 
      \multicolumn{3}{|c|}{\textbf{APA}} \\
      \cline{3-14}
      & \textbf{\scriptsize Demographics}&  P & R& $F_1$& P& R& $F_1$& P& R& $F_1$& P& R& $F_1$\\
    \hline
         bioelectricity-2& 249 (3.01)& 0.95& 0.54& 0.69& 0.89& 0.65& 0.75& 0.81& \textbf{0.81}& \textbf{0.81}& 0.91& 0.57& 0.70\\
         calc-1& 965 (1.52)& 0.83& 0.57& 0.67& 0.85& 0.52& 0.65& 0.80& 0.63& 0.71& 0.90& 0.48& 0.63\\
         bioinfo1-1& 234 (1.52)& 0.67& 0.46& 0.55& 0.70& \textbf{0.46}& \textbf{0.56}& 0.67& 0.46& 0.55& 0.69& 0.42& 0.52\\
         maththink-004& 494 (0.94)& 0.56& 0.47& 0.51& 0.63& 0.55& 0.59& 0.64& \textbf{0.57}& \textbf{0.60}& 0.55& 0.49& 0.52\\
         comparch-2& 132 (0.86)& 1.0& 0.62& \textbf{0.76}& 1.00& 0.62& \textbf{0.76}& 0.71& \textbf{0.77}& 0.74& 1.00& 0.54& 0.70\\
         ml-5& 2058 (0.81)& 0.74& 0.63& 0.68& 0.74& 0.65& \textbf{0.69}& 0.65& \textbf{0.71}& 0.68& 0.74& 0.65& \textbf{0.69}\\
         rprog-3& 1123 (0.49)& 0.51& 0.63& 0.56& 0.57& 0.57& 0.57& 0.52& \textbf{0.70}& \textbf{0.60}& 0.53& 0.64& 0.58\\
         casebased-2& 121 (0.26)& 0.50& \textbf{0.83}& \textbf{0.63}& 0.44& 0.67& 0.53& 0.33& 0.67& 0.44& 0.45& 0.83& 0.59\\
         gametheory2-1& 122 (0.22)& 0.08& 0.25& 0.13& 0.60& \textbf{0.75}& \textbf{0.67}& 0.38& 0.75& 0.50& 0.09& 0.25& 0.13\\
         smac-1& 618 (0.21)& 0.24& 0.65& 0.35& 0.36& \textbf{0.53}& \textbf{0.43}& 0.26& 0.53& 0.35& 0.28& 0.76& 0.41\\
         medicalneuro-2& 323 (0.09)& 0.20& 0.60& 0.30& 0.30& 0.60& \textbf{0.40}& 0.25& 0.40& 0.31& 0.13& 0.60& 0.21\\
         compilers-4& 616 (0.02)& 0.0& 0.0& 0.0& 0.0& 0.0& 0.0& 0.0& 0.0& 0.0& 0.0& 0.0& 0.0\\
         \hline
         Macro Avg.& & 0.52& 0.52& 0.52& \textbf{0.59}& 0.55& \textbf{0.57}& 0.50& \textbf{0.58}& 0.54& 0.52& 0.52& 0.52\\
         Weighted & & 0.56& 0.54& 0.55& \textbf{0.60}& 0.54& \textbf{0.57}& 0.54& \textbf{0.59}& 0.56& 0.58& 0.55& 0.56\\
         Macro Avg& & & & & & & & & & & & & \\
         \hline
    \end{tabular}
    \caption{Prediction performance for the positive class (intervened threads) of the three proposed models versus a neural baseline hLSTM, using hierarchical encoding of the discussion thread. Best performance in $F_1$ and Recall is bolded. Column 2 shows \# of threads in each course and the intervention ratio, the ratio of intervened to non-intervened threads, in parentheses.}
    \label{tab:result_attn_all}
\end{table*}

\subsection{Results}
\label{ss:results}
% Min NAACL: the table doesn't make a lot of sense to me.  You compare with the baseline of EDM, but this is weaker by your own assessments that hLSTM with full context.  Shouldn't you propose that as the comparative one and then show the microanalysis against that one? 
Table~\ref{tab:result_attn_all} shows performance of all our proposed models and the  
neural baseline over our 12 MOOC dataset.  
Our models of UPA, PPA individually better the baseline by 5 and 2\% 
on $F_1$ and 3 and 6\% on recall respectively. 
UPA performs the best in terms of $F_1$ on average 
while PPA performs the best in terms of recall on average. At the individual course level, 
however, the results are mixed. UPA performs the best on $F_1$ on 5 out of 12 courses, PPA 
on 3 out 12 courses, APA 1 out of 12 courses and the baseline hLSTM on 1. 
PPA performs the best on recall on 7 out of the 12 courses.
We also note that course level performance differences correlate with the course size 
and intervention ratio (hereafter, \textit{i.ratio}), which is the ratio of intervened to non-intervened threads. UPA performs better than PPA and APA on low intervention courses 
(i.ratio $\lessapprox$ 0.25) mainly because PPA and APA's performance drops steeply 
when i.ratio drops (see col.2 parenthesis and $F_1$ of PPA and APA). 
While all the proposed models beat the baseline on every course 
except casebased-2.
On medicalneuro-2 and compilers-4 which have the lowest 
i.ratio among the 12 courses none of the neural models better 
the reported baseline~\cite{chandrasekaran2015learning} (course level 
not scores not shown in this paper). The effect is pronounced in compilers-4 
course where none of the neural models were able to predict any intervened 
threads. This is due to the inherent weakness of standard neural models, which are unable 
to learn features well enough when faced with sparse data.

The best performance of UPA indicates that the reply context of the instructor's post 
$P_n$ correlates strongly with that of the previous post $P_{n-1}$. 
% Min: repetitive, you said this before.
This is not surprising 
since normal conversations are typically structured that way.

\begin{table}[]
    \centering
    \begin{tabular}{|l|r|r|r|}
        \hline
        \textbf{Baseline Model} & \textbf{P}& \textbf{R}& $\mathbf{F_1}$\\
        \hline
        \small \citeauthor{chandrasekaran2015learning}             & & &\\
        \small \citeyear{chandrasekaran2015learning} (non-neural)  & 0.47& 0.47& 0.47\\
        \hline
        hLSTM  & & & \\
        \small with context = $<P_1,..,P_{n-1}>$ & 0.52& 0.52& 0.52\\
        \hline
        hLSTM  & & &\\
        \small with single post context = $P_{n-1}$  & 0.37& 0.44& 0.41\\
        %\hline
        %CNN-LSTM  & & &\\
        \hline
        
    \end{tabular}
    \caption{Comparison across baseline models.}
    \label{tab:baselines}
\end{table}

%%%%%%%%%%%%%%%%%%%%%%%%%%%%%%%%%%%%%%%%%%%%%%%%%%%%%%%%%%%%%%%%%%%%%
\section{Discussion}
\label{s:discuss}
In order to further understand the models' ability to infer the context and 
its effect on intervention prediction, we further investigate the following research 
questions.~\\
\textbf{RQ1}. {\it Does context inference help intervention prediction? }~\\
In order to understand if context inference is useful to intervention 
prediction, we ablate the attention components and experiment with  
the vanilla hierarchical LSTM model. Row~3 of Table~\ref{tab:baselines} shows 
the macro averaged result from this experiment. The UPA and PPA attention models 
better the vanilla hLSTM by 5\% and 2\% on average in $F_1$ respectively. Recall  
that the vanilla hLSTM already has access to a context consisting of all 
posts (from $P_1$ through $P_{n-1}$). In contrast, the UPA and PPA models selectively infers  
a context for $P_{n-1}$ and $P_{n-2}$ posts, respectively, and use it to predict
intervention. The improved performance of our attention models 
that actively select their optimal context, over a model with the complete thread as context, hLSTM, 
shows that the context inference improves intervention prediction over using the 
default full context.~\\
%possible example with inferred context

\begin{figure*}[t!]
    \centering
    \begin{subfigure}[t]{0.33\textwidth}
        %\centering
        \includegraphics[height=1.2in]{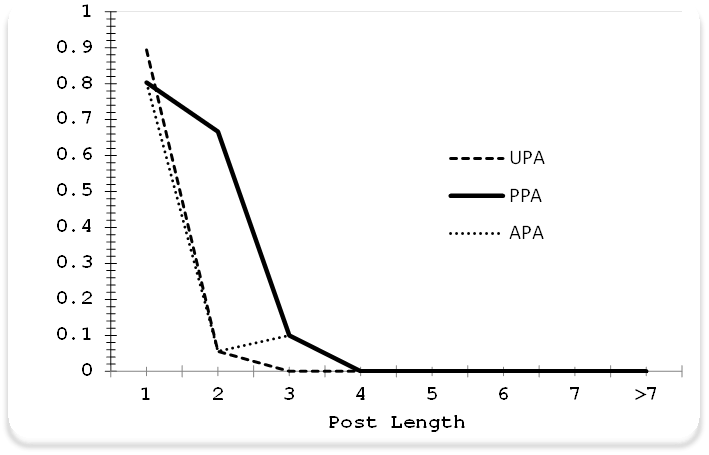}
        \caption{Recall for course calc1-3}
    \end{subfigure}%
     ~
    \begin{subfigure}[t]{0.33\textwidth}
        %\centering
        \includegraphics[height=1.2in]{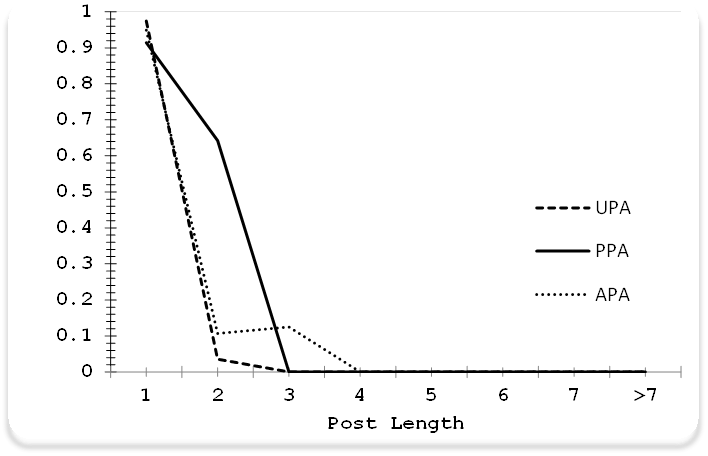}
        \caption{Recall for course ml-5}
    \end{subfigure}%
    ~
    \begin{subfigure}[t]{0.33\textwidth}
        %\centering
        \includegraphics[height=1.2in]{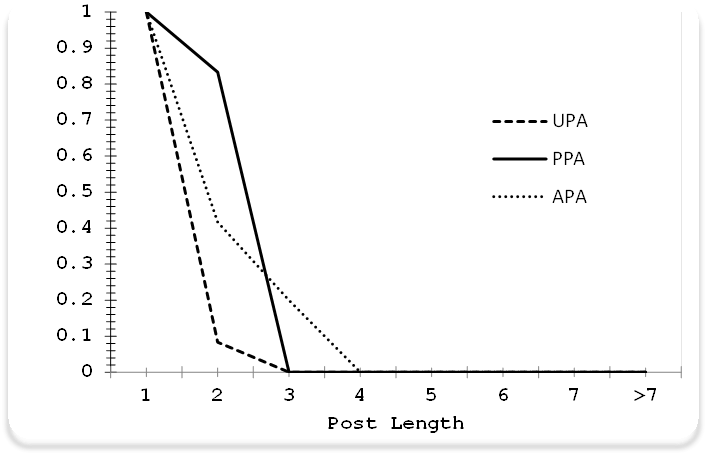}
        \caption{Recall on course rprog-3}
    \end{subfigure}
    \caption{Recall scores plotted across different thread lengths. UPA's performance drops 
    steeply on threads of length $>$ 1. Although APA's performance on short threads is 
    worse than others, it is better at predicting long threads which is a key objective 
    of the model.}
    \label{fig:lengthanalysis}
\end{figure*}

\begin{figure}[h]
    \centering
    \includegraphics[width=0.5\textwidth]{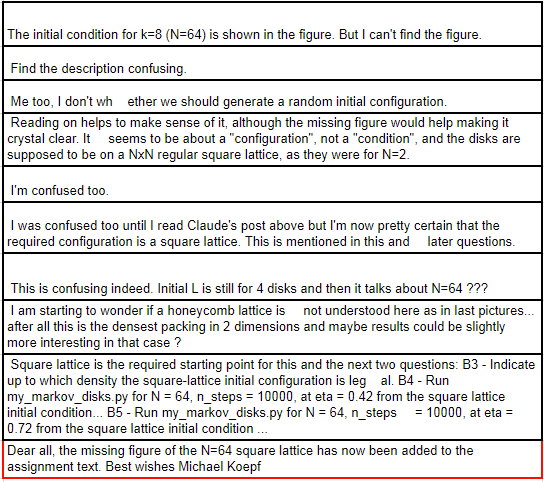}
    \caption{An example of a long thread from smac-1 with nine 
    posts that the APA model accurately predicted but not by UPA and PPA.
    }
    \label{fig:longpost}
\end{figure}

\textbf{RQ2}. {\it How well do the models perform across threads of different lengths? }
To understand the models' prediction performance across threads of different 
lengths, we bin threads by length and study the models' recall. 
We choose three courses, ml-5, rprog-3 and calc-1, 
from our corpus of 12 with the highest number of positive 
instances ($>$100 threads). We limit our analysis to these since binning renders 
courses with fewer positive instances sparse. Figure~\ref{fig:lengthanalysis} shows 
performance across thread lengths from 1 through 7 posts and $>7$ posts.
Clearly, the UPA model performs much better on shorter threads than on longer threads 
while PPA and APA works better on longer threads. 
Although, UPA is the best performing model in terms of overall $F_1$
its performance drops steeply on threads of length $>1$. 
UPA's overall best performance is because most of the interventions in the 
corpus happen after one post.
To highlight the performance of APA we show an example from smac-1 in Figure~\ref{fig:longpost} with nine posts which was predicted correctly as 
intervened by APA but not by other models. Threads shows students confused 
over a missing figure in a homework. The instructor finally shows up, though late, 
to resolve the confusion.

~\\

\textbf{RQ3}. {\it Do models trained with different context lengths perform better than 
when trained on a single context length? }~\\
We find that context length has a regularising effect on the model's performance 
at test time. This is not surprising since models trained with threads of single 
context length will not generalise to infer different context lengths. 
Row~4 of Table~\ref{tab:baselines} shows a steep performance drop in training by  
classifier with all threads truncated to a context of just one post, $P_{n-1}$, 
the post immediately preceding the intervened post. 
We also conducted an experiment with a multi-objective loss function 
with an additive cross-entropy term where each term computes loss from 
a model with context limited to a length of 3.
We chose 3 since intervened threads in all the courses had a median 
length between 3 and 4. We achieved an $F_1$ of 0.45 with a precision of 
0.47 and recall of 0.43. This achieves a performance comparable to 
that of the \cite{chandrasekaran2015learning} with context 
length set to only to 3. This approach of using infinitely 
many loss terms for each context length from 1 through the maximum 
thread length in a course is naive and not practical. We only 
use this model to show the importance of training the model 
with loss from threads of different lengths to prevent 
models overfitting to threads of specific context lengths.
~\\

%\textbf{RQ3}. Does the attended context representation contain intervention triggers?

%\textbf{RQ}. How do the models' precision-recall trade-off affect predicting threads that were missed

%\textbf{Limitations}.

%%%%%%%%%%%%%%%%%%%%%%%%%%%%%%%%%%%%%%%%%%%%%%%%%%%%%%%%%%%%%%%%%%%%%
\section{Conclusion}
\label{s:conc}
We predict instructor intervention on student discussions by 
first inferring the optimal size of the context 
needed to decide on the intervention decision for the intervened 
post. We first show that a structured representation of the 
complete thread as the context is better than a bag-of-words, 
feature-rich representation. We then propose attention-based 
models to infer 
% Min (via Slack) NAACL
% contexts are always contiguous sets of posts, correct?
% Do you also validate your claim about early and late intervention modeling?
and select a context -- defined as a contiguous subsequence of 
student posts -- to improve over a model that always 
takes the complete thread as a context to prediction intervention. 
Our Any Post Attention (APA) model enables instructors to tune the 
model to predict intervention early or late.  We posit our APA model 
will enable MOOC instructors employing  
varying pedagogical styles to use the model equally well. 
We introspect the attention
models' performance across threads of varying lengths and show 
that APA predicts intervention on longer threads, which possesses 
more candidate contexts, better.

%future work
We note that the recall of the predictive models for longer threads 
(that is, threads of length greater 2) can still be improved. 
Models perform differently between shorter and longer length. 
An ensemble model or a multi-objective 
loss function is thus planned in our future work to better 
prediction on such longer threads.

%\section*{Acknowledgments}
%The authors would like to thank Shamil Chollampatt 
%for reviewing the paper, feedback, review of the models 
%and help with implementation.
%This research is funded in part by NUS Learning Innovation Fund --
%Technology grant \#C-252-000-123-001, and by the Singapore National
%Research Foundation under its International Research Centre @
%Singapore Funding Initiative and administered by the IDM Programme
%Office.
%We thank NUS Centre for Instructional Technology, Andreina Parisi-Amon 
%from Coursera and Prof. Bernard Tan for helping us acquire legal 
%permission to use Coursera's data for our academic research. 

\bibliography{acl2019}
\bibliographystyle{acl_natbib}
%\appendix

\end{document}